%% file: main.tex
\documentclass[10pt,twocolumn,letterpaper]{article}

\usepackage[pagenumbers]{cvpr} 

\usepackage{graphicx}
\usepackage{array}
\usepackage{multirow}
\usepackage{amsmath}
\usepackage{amssymb}
\usepackage{textcomp}  
\usepackage{gensymb}
\usepackage{booktabs}

\usepackage{xcolor}  
\usepackage{siunitx}  
\usepackage{appendix}

\usepackage{microtype}

\usepackage{pifont}
\definecolor{citecolor}{HTML}{65C3A6}
\usepackage[pagebackref,breaklinks,colorlinks,citecolor=citecolor]{hyperref}
\usepackage{colortbl}

\usepackage[capitalize]{cleveref}
\crefname{section}{Sec.}{Secs.}
\Crefname{section}{Section}{Sections}
\Crefname{table}{Table}{Tables}
\crefname{table}{Tab.}{Tabs.}

\newif\ifsubmit
\submittrue

\newcommand{\cutsectionup}{\vspace*{-2pt}}

\newcommand{\cuthalfcaptionup}{\vspace*{-5pt}}

\newcommand{\cutabstractup}{\vspace*{-10pt}}

\begin{document}

\input{macros.tex}

\title{Mip-NeRF RGB-D: Depth Assisted Fast Neural Radiance Fields}

\author{
Arnab Dey
\qquad
Yassine Ahmine
\qquad
Andrew I. Comport
\\[9pt]
I3S-CNRS/Universit\'e C\^ote d'Azur\\
Sophia-Antipolis, France \\[1mm]
}

\twocolumn[{%
\renewcommand\twocolumn[1][]{#1}%
\maketitle
\vspace{-18pt}
\input{figs_tex/teaser}
}]

\input{content_main}

{\small
\bibliographystyle{ieee_fullname}
\bibliography{reference}
}

\newpage 
\onecolumn
\section*{Supplementary Material: \\Mip-NeRF RGB-D: Depth Assisted Fast Neural Radiance Fields}
\vspace{0.5\baselineskip}
\par In this supplementary material, some important definitions of terms will be presented, and finally more quantitative and qualitative results on the real RGB-D datasets will be provided.  

\subsection*{Definitions}
\subsubsection*{Depth uncertainty:}
For the purpose of this paper, the sensor depth uncertainty is assumed to be Gaussian in inverse depth as $ \frac{1}{D}~\sim~\mathcal{N}(\mu,\sigma)$ (i.e. the probability of the measured depth from its true value).
\subsection*{The network architecture}
\begin{figure*}[ht!]
\begin{center}
  \includegraphics[width=0.8\linewidth]{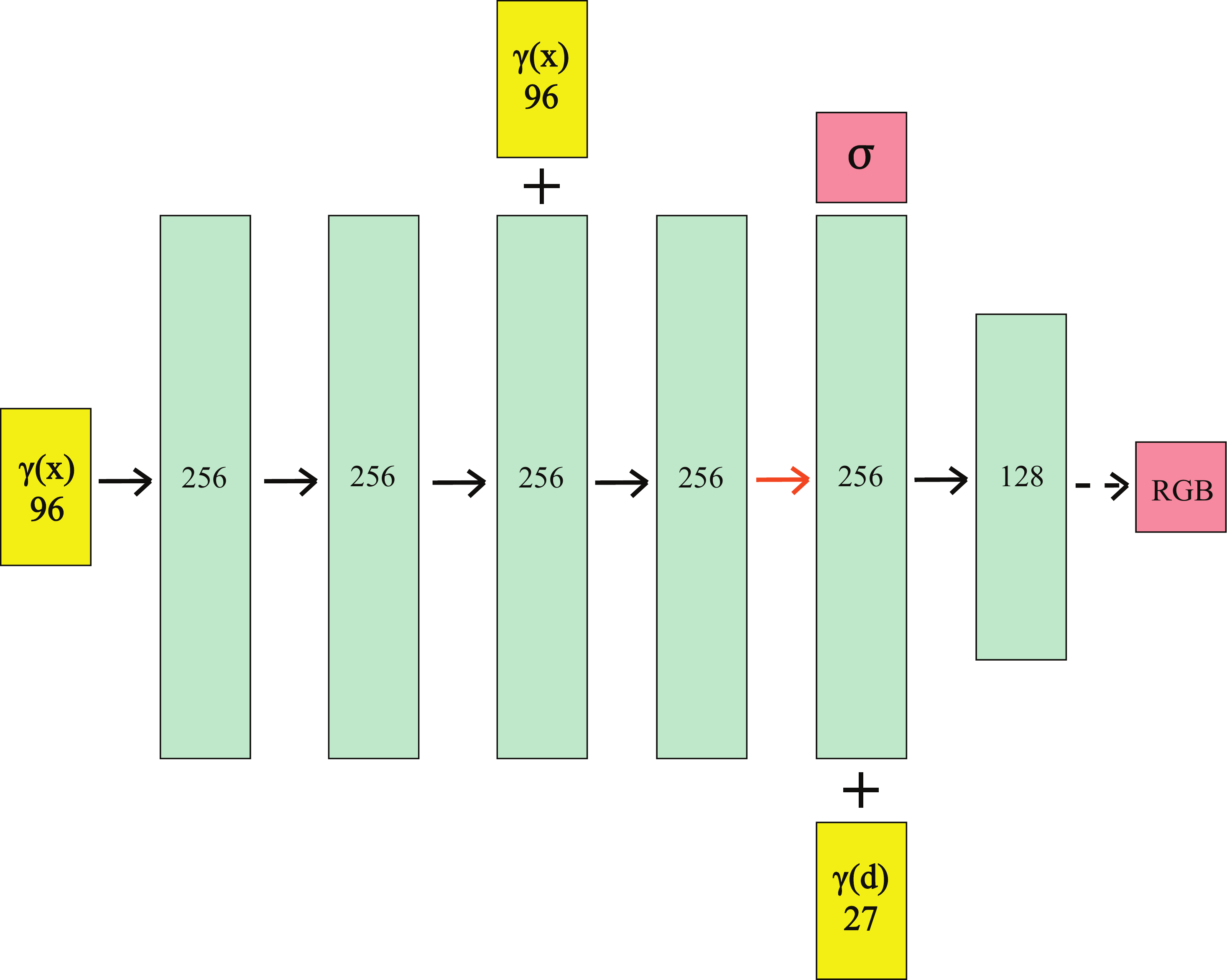}
\end{center}
  \caption{\textbf{Proposed network architecture:} A visualization of the fully-connected multilayer perceptron network. Encoded input vectors are shown in yellow, hidden layers are shown in green, output vectors are shown in red, the numbers inside the blocks represent the size of the vector. The network uses fully connected layers, ReLU activation layers are represented by black arrows, no activation layers with orange arrows, "$+$" represents vector concatenation and Sigmoid activation is represented by dashed arrows. Positional encoding of input point x is denoted by $\gamma(x)$ and direction encoding by $\gamma(d)$. The network uses 4 fully-connected ReLU layers with 256 channels. The network architecture is similar to the NeRF~\cite{mildenhall2020nerf} architecture but smaller. A fully connected layers with 128 channels is use with Sigmoid activation to output RGB radiance corresponding to the input point.}
\label{fig:network}
\end{figure*}

As mentioned before, the proposed method uses a smaller but more efficient network architecture compared to NeRF\cite{mildenhall2020nerf}. The MLP uses 4 hidden layers with ReLU activation; the viewing direction is added to the last layers before outputting the RGB values. The network architecture is shown in detail in Figure:\ref{fig:network}.

\subsection*{Additional results}
The proposed method has been extensively tested on different simulated datasets and this section presents three further experiments that have been performed on noisy datasets. Prior to testing using real depth images, the effect of noisy depth measurements was studied by adding a Gaussian noise with a standard deviation of $0.01$ m to the inverse depth. The objective here was to simulate the uncertainty inherent in depth measurements. The results of this experiment are shown in Table~\ref{table:noise}. The result shows that, despite the added noise, the proposed method is able to maintain its quality and produce better results than DONeRF. 

\begin{table}[hb]
\small
	\centering
	\begin{tabular}{p{1.2cm}|p{0.9cm} p{0.9cm} p{1.2cm} p{0.9cm}}
	\hline
      & \multicolumn{4}{|c}{Metrics} \\
    \hline
    Method & PSNR$\uparrow$ & SSIM$\uparrow$ & AbsRel$\downarrow$ & LPIPS$\downarrow$\\ 
	\hline
	DONeRF & 30.91 & 0.94 & 0.03 & 0.0009 \\
	\textbf{Proposed} & \cellcolor[HTML]{32CB00}32.82 & 0.94 & \cellcolor[HTML]{32CB00}0.01 & \cellcolor[HTML]{32CB00}0.0004\\
	\end{tabular}
	\caption{Quantitative comparison between DONeRF and the proposed method when depth noise is added to lego dataset.}
	\label{table:noise}
\end{table}

\clearpage


\end{document}

%% file: macros.tex
\definecolor{yellow}{rgb}{1,1, 0.6}
\definecolor{lightyellow}{rgb}{1,1, 0.8}
\definecolor{orange}{rgb}{1, 0.8, 0.6}
\definecolor{red}{rgb}{1, 0.6, 0.6}

\definecolor{wincolor}{rgb}{0.85, 0.0, 0.0}

\definecolor{darkyellow}{rgb}{0.8, 0.8, 0.5}
\definecolor{darkred}{rgb}{0.7, 0.3, 0.3}
\definecolor{darkgreen}{rgb}{0.3, 0.7, 0.3}
\definecolor{blue}{rgb}{0, 0, 1.0}
\definecolor{green}{rgb}{0, 1.0, 0}
\definecolor{pink}{rgb}{1, 0.4, 0.7}

\newcommand{\matt}[1]{{\color{blue} Matt: #1}}

\newcommand{\modeltheta}{\mathrm{\Theta}}
\newcommand{\absrp}{\sigma}

\newcommand{\numsamples}{N}
\newcommand{\numsamplescoarse}{N_c}
\newcommand{\numsamplesfine}{N_f}

\newcommand{\posxy}{xy}
\newcommand{\posxyz}{xyz}
\newcommand{\angletheta}{\theta}
\newcommand{\anglephi}{\phi}
\newcommand{\posall}{\posxyz\angletheta\anglephi}

%
%

%% file: figs_tex/teaser.tex
\begin{center}
\captionsetup{type=figure}
\cuthalfcaptionup
    \includegraphics[width=\linewidth]{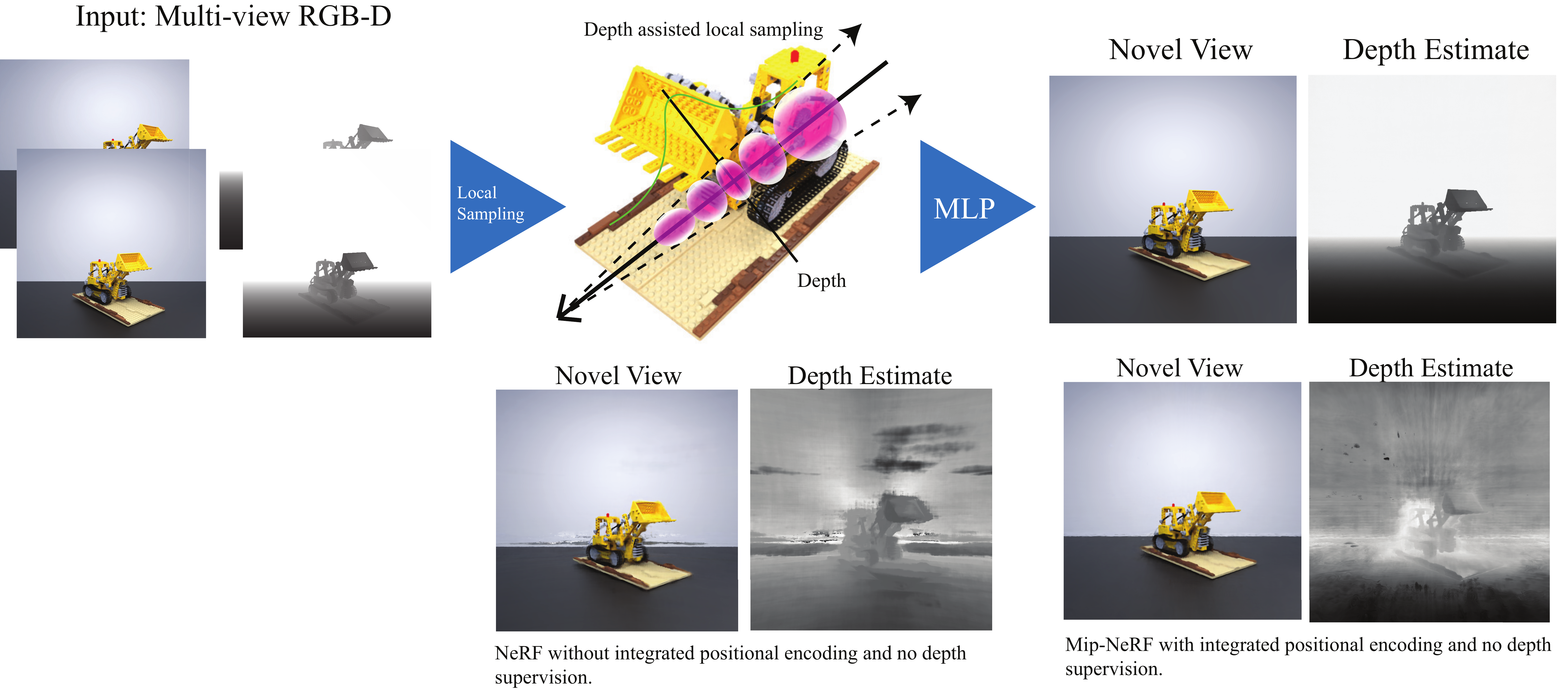}
    \vspace{-20pt}
\captionof{figure}{Mip-NeRF RGB-D uses RGB-D frames to represent 3D scenes using neural radiance fields. Depth information is used for local sampling and geometric loss. It produces significantly better photometry and geometry.}
\cuthalfcaptionup
\label{fig:teaser}
\end{center}%

%% file: content_main.tex
\begin{abstract}
Neural scene representations, such as Neural Radiance Fields (NeRF), are based on training a multilayer perceptron (MLP) using a set of color images with known poses. An increasing number of devices now produce RGB-D(color + depth) information, which has been shown to be very important for a wide range of tasks. Therefore, the aim of this paper is to investigate what improvements can be made to these promising implicit representations by incorporating depth information with the color images. In particular, the recently proposed Mip-NeRF approach, which uses conical frustums instead of rays for volume rendering, allows one to account for the varying area of a pixel with distance from the camera center. The proposed method additionally models depth uncertainty. This allows to address major limitations of NeRF-based approaches including improving the accuracy of geometry, reduced artifacts, faster training time, and shortened prediction time. Experiments are performed on well-known benchmark scenes, and comparisons show improved accuracy in scene geometry and photometric reconstruction, while reducing the training time by 3 - 5 times.

\end{abstract}

\cutabstractup

\cutsectionup
\section{Introduction}
Recent advances in neural scene representations \cite{sitzmann2019scene, mildenhall2020nerf} have demonstrated that neural networks can be used to represent 3D scenes as weights of a neural network for the purpose of rendering novel photorealistic views. Methods such as \cite{mildenhall2020nerf, saito2019pifu, lombardi2019neural} learned a volumetric representation from a sparse set of RGB images captured from color camera sensors. This method requires precomputation of camera poses and uses two multilayer perceptron networks to represent scene geometry and lighting effects. Although the NeRF models and their variants have shown impressive results, the underlying model is computationally inefficient, largely due to its volumetric search space for intersecting viewing rays, leading to extended training times. For example, volume rendering involves sampling points along each viewing ray (256 for NeRF) to calculate the color of the ray from the volume density and radiance of each sample point. Furthermore, the multiview triangulation problem is sometimes intractable from only images, which leads to artifacts and inaccurate geometry. 

Although color-only approaches work well for applications that only have RGB images available, this approach can be improved by considering depth information alongside color. Many devices, including mobile phones, now include RGB-D sensors, and the aim of this paper is to investigate and devise a methodology to incorporate depth information into a neural scene representation. 

Only a few methods have been proposed to take advantage of depth measurements simultaneously with color within the volumetric rendering pipeline~\cite{neff2021donerf,kangle2021dsnerf}. However, these methods do not explicitly model the uncertainty of the sensor. To successfully incorporate noisy depth measurements into the volumetric rendering pipeline, the recent Mip-NeRF approach~\cite{barron2021mip} provides a framework that accounts for the uncertainty of color pixel with varying depth by replacing classic 3D ray sampling with conic region sampling. This approach provides an elegant framework for including multivariate Gaussian uncertainty and will be extended in this paper to include depth uncertainty.

In this paper, it will be demonstrated that considering depth information can improve geometry considerably compared to only color information at several different levels. First, this method shows how local sampling along the rays, guided by surface information from RGB-D frames, can reduce the number of samples along the ray and replace the coarse network of NeRF. Second, a joint color-and-depth-loss term will be shown to allow the network to learn the geometry and color of the scene from a limited number of input views. Third, the proposed method shows how depth uncertainty can be incorporated into a multivariate Gaussian method to query the MLP. Finally, an adaptive training method will be proposed that allows the network to learn multiple scales of uncertainty within the representation.

\par To sum up, the proposed method is based on a RGB-D neural radiance field combined with an implicit occupancy representation that takes into account both color and depth observations. 
\newline The key contributions of the article are summarized as follows:
\begin{itemize}
    \item Depth information is used for efficient sampling. 
    \item The representation is optimized simultaneously on scene geometry and photometry.
    \item Depth uncertainty is handled adaptatively via a new local sampling strategy.
\end{itemize}

\begin{figure*}[t]
\begin{center}
   \includegraphics[width=1\linewidth]{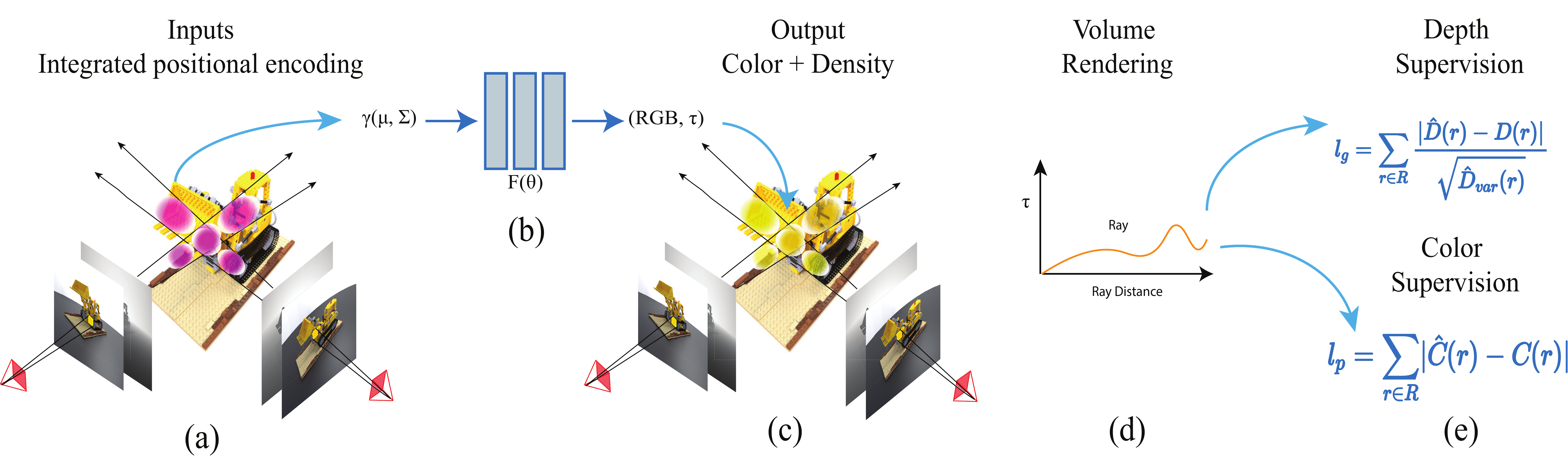}
\end{center}
   \caption{An overview of the proposed Mip-NeRF RGB-D. \textbf{(a)} The input to the network is the integrated positional encoding of a conical frustum segment. \textbf{(b)} The network outputs volume density and color. \textbf{(c)} The color and depth of a ray is generated using the classic volume rendering method. \textbf{(d)} The network is optimized using a color and depth loss.}
\label{fig:method}
\end{figure*}

\cutsectionup
\section{Related Work}

The proposed Mip-NeRF RGB-D uses a set of RGB-D inputs to learn a volumetric scene representation of the observed scene using a multilayer perceptron by leveraging both depth and color information. In the following, related work to this research will be discussed.
\subsection{\emph{Novel view synthesis from images}} Image-based view synthesis uses a number of techniques to generate novel images, such as transforming or warping an existing set of images using estimated geometry and camera poses to create novel views \cite{hedman2016scalable, gortler1996lumigraph}. \cite{heigl1999plenoptic} used a sequence of images and directly rendered the views by projective mapping of all images to a common plane of mean geometry. To generate a novel view from a set of captured images of different poses requires blending them to target views; even though the geometry of the static objects is constant in different views, the appearance can change depending on lighting and object properties. To overcome these drawbacks \cite{hedman2018deep,thies2020image} used artificial neural networks to reduce artifacts and view-dependent effects in the generated novel views. 
\subsection{\emph{Implicit neural surface representation}}
These methods use neural networks to learn a neural surface representation of the object using voxels, meshes, and point cloud data. Although they are capable of achieving impressive results, they are limited by their internal resolution and high-frequency details. Mescheder et al. \cite{mescheder2019occupancy} used a neural network to learn a continuous 3D occupancy function; Given 3D points as input to an occupancy network, the network predicts binary occupancy at that 3D location. Later, \cite{chen2019learning} used an MLP to predict occupancy from a feature vector and the 3D coordinates of the location. On the other hand, \cite{park2019deepsdf} learned a signed distance function(SDF) instead of occupancy to improve the quality of the reconstruction. \cite{saito2019pifu} showed that it is possible to infer 3D surfaces and texture from a single image using an implicit function.
\subsection{\emph{Neural volume rendering}}
 Lombardi et al. \cite{lombardi2019neural} initially introduced volume rendering for novel view synthesis using a CNN-based encoder and an MLP-based decoder to produce density and color for each point in space. The well-known Neural Radiance Fields approach was introduced in \cite{mildenhall2020nerf} and demonstrated compelling results with a simple method that takes 3D points and the associated view direction as input to an MLP and outputs density and color. Some of the drawbacks of NeRF are its long training time, its long rendering time, the need to train a separate model for each scene, and it only works on static scenes. Various investigations have been conducted since the original NeRF to address these problems. \cite{Liu20neurips_sparse_nerf, neff2021donerf, Garbin21arxiv_FastNeRF, reiser2021kilonerf, yu2021plenoctrees} addressed the slow inference time of NeRF by using a tiny MLP and a better sampling strategy. \cite{kangle2021dsnerf} used sparse depth supervision during training to improve the training time of the NeRF model. \cite{Pumarola20arxiv_D_NeRF, Gafni20arxiv_DNRF, noguchi2021neural} address the problem of static scenes. \cite{Yu20arxiv_pixelNeRF, Schwarz20neurips_graf, Tancik20arxiv_meta, Chan20arxiv_piGAN} have generalized NeRF models using fully convolutional image features, a generator discriminator, and meta-learning. Saito et al.\cite{barron2021mip} focused on NeRF aliasing and sampling problems. They proposed an integrated positional encoding, which uses a conical frustum defined by the mean and covariance of the rays, and the neural radiance field is integrated over the region represented by 3D Gaussian encoding.  

\subsection{\emph{Neural radiance field with depth}}
To solve the problem of incorrect geometry prediction when a limited number of input views are given, \cite{kangle2021dsnerf} proposed using depth as alternate supervision. Ds-NeRF uses a sparse 3D point cloud and then reprojects the errors between the detected 2D keypoints and projected 3D points, generated by commonly used structure-from-motion (SfM) algorithms which are error-prone. They optimize the model over a combined color and depth loss function.
 Similarly, NerfingMVS \cite{wei2021nerfingmvs} uses a monocular depth network to generate depth prior from SfM reconstruction of the scene. The adapted depth priors are used to guide the sampling process of points along the ray. Unlike DS-NeRF, it generates a dense depth prior from sparse SfM points using a pretrained depth network. Azinovic et al.\cite{azinovic2021neural} also demonstrated the incorporation of depth with NeRF to produce a better and more detailed reconstruction than simply using color or depth alone. Unlike others, it uses a truncated signed distance function(TSDF) instead of volume density to represent the underlying geometry. It still uses two networks that significantly affect training and prediction time.
On the other hand, iMap \cite{sucar2021imap} shows that NeRF can be used to represent scenes in a real-time SLAM system. It jointly optimizes the 3D map and camera pose using keyframes. iMap uses a smaller MLP (4 layers) than NeRF and does not consider the viewing direction to model lighting effects. 
DONeRF \cite{neff2021donerf} proposed a compact dual network design to reduce evaluation cost, leading to a faster prediction time. The coarse NeRF network is replaced by a \textit{depth oracle network} based on a classification network. To reduce the number of samples along the rays, they suggested nonlinear transformation and a local sampling strategy, which helped them to achieve a similar result to NeRF with a fraction of the samples, but the method is limited to forward facing scenes.

\cutsectionup
\section{Method}
\input{method}

\cutsectionup
\section{Experimental Results} 
 In this section, the proposed methods are evaluated on various datasets and compared with other state-of-the-art NeRF based methods. 
 \subsection{Experimental Setup} 
 \subsubsection{Datasets}
  Simulated datasets were used for all experiments. Each data set contained RGB images, depth maps, and their corresponding camera poses. All poses in the datasets belong to an upper hemisphere, where the object is placed in the center. Four different scenes were considered for the experiments: Lego\footnote{\url{https://www.blendswap.com/blend/11490}}, Cube, Human \footnote{\url{https://renderpeople.com/free-3d-people/}}, and Drums\footnote{\url{https://www.blendswap.com/blend/13383}}. The input images have $800\times800$ resolution, and the depth measurements are in meters. Each of the datasets has three versions, in which the number of training images is 8, 30, and 100.
 
 \subsubsection{Implementation Details}
 The proposed method is implemented using a combination of PyTorch and CUDA. The ADAM optimizer with a learning rate of $5\times10^{-4}$ and an exponential decay of the learning rate of $5\times10^{-1}$ in every five epoches has been used. A batch size of 2048 on 2 Nvidia Rtx 3090 GPUs was used for all experiments. 16 frequency bands were used for integrated positional encoding of the conical frustum and 4 frequency bands to encode viewing directions with positional encoding. \par
 For all experiments, the following parameters are used as default: the photometric scale factor for the loss function is $ \lambda_p = 100 $, standard deviation for Gaussian sampling is 0.3, \(\lambda_r=0.09\) and \( \lambda_m = 0.1\) for adaptive sampling.  

 \subsubsection{Metrics}
  Four metrics are used to evaluate the predicted RGB image quality and depth map: \textit{Peak Signal-to-Noise Ratio (PSNR in dB}): to compare the quality of the RGB reconstruction, the higher is better; \textit{Absolute Relative distance} (Abs Rel in m): to compare the quality of the generated depth map, the lower is better; \textit{Structural Similarity Index} (SSIM in \%) \cite{wang2004image}: quantifies the degradation of image quality in the reconstructed image, the higher is better; \textit{Learned Perceptual Image Patch Similarity} (LPIPS) \cite{zhang2018perceptual}: the distance between the patches of the image, the lower means that the patches are more similar.
  
\subsection{Comparison}
First, local sampling strategies are compared in Section \ref{localsampling}. Then, the effect of a different number of samples on the proposed model is discussed. After that, the proposed method is applied in different scenes. Finally, in Section \ref{comparenerf}, the proposed method is compared with other NeRF-based methods that use depth supervision.

\subsubsection{Comparison between different local sampling strategies}\label{localsampling}
Instead of placing samples over the entire ray, local sampling places samples only on the relevant regions of the ray using depth information. In this section, proposed local sampling strategies are compared.  Table \ref{table:sampling} shows the quantitative performance of the three sampling strategies mentioned. 

\begin{table}[htb]
	\centering
	\begin{tabular}{p{1.5cm}|p{1cm} p{1cm} p{1cm} p{1cm}}
	\hline
      & \multicolumn{4}{|c}{Metrics} \\
    \hline
    Strategy & PSNR $\uparrow$ & SSIM $\uparrow$ & Abs Rel$\downarrow$ & LPIPS $\downarrow$ \\ 
	\hline
	Equidistant & 19.86 & 0.87 & \cellcolor[HTML]{32CB00} 0.02 & 0.0023 \\

	Gaussian & 20.75 & 0.88 & 0.04 & \cellcolor[HTML]{32CB00}0.0022 \\
	
	\textbf{Adaptive} & \cellcolor[HTML]{32CB00} 21.09 & \cellcolor[HTML]{32CB00} 0.89 & 0.04 & \cellcolor[HTML]{34FF34}0.0024 \\
	
	NeRF & 19.21 & 0.88 & \cellcolor[HTML]{CB0000}0.34 & 0.0036 \\
	
	\end{tabular}
	
	\caption{Comparison of three different sampling strategies. The Lego spherical dataset containing 8 training images has been used for all experiments.  All experiments used 16 sampling points per ray. Best values are highlighted by green, significant wrose values by red, and darker shades represent best values.}
	\label{table:sampling}
\end{table}

The results show that adaptive sampling performs best among the proposed sampling strategies. All local sampling strategies improve the underlying geometry compared to NeRF. 
\begin{figure*}[ht!]
\begin{center}
   \includegraphics[width=0.65\linewidth]{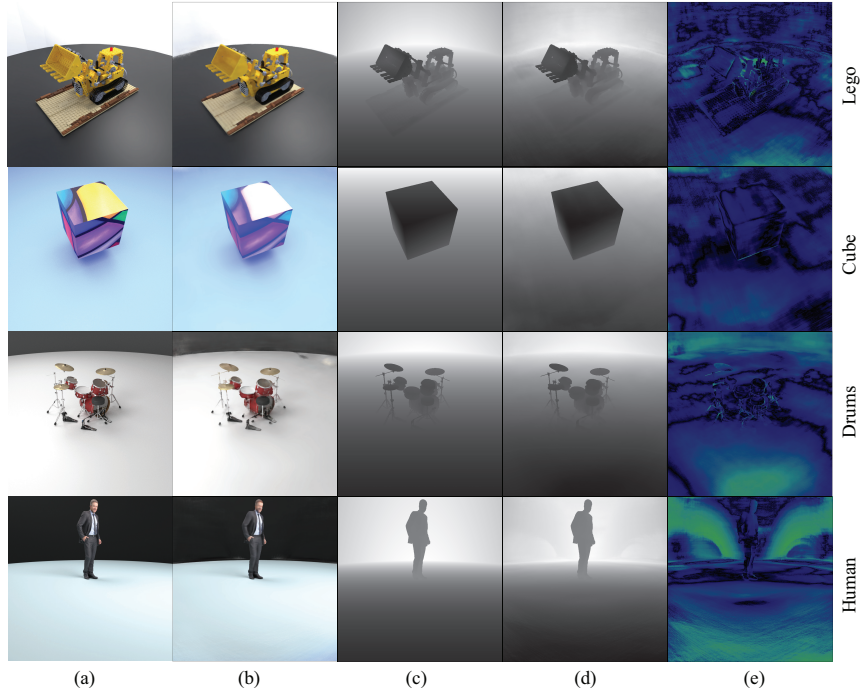}
\end{center}
   \captionof{figure}{Qualitative comparison on blender scenes: Visual comparison between generated RGB ground truth images and the true depth maps generated by the proposed method, where (a) Ground truth RGB images; (b) Predicted RGB images; (c) True depth maps; (d) Predicted depth maps; (e) Absolute error between predicted and true depth maps.}
\label{fig:lcd}
\end{figure*}
\subsubsection{Effect of ray sample size} 
NeRF based methods use a large number of ray samples to estimate volume density and produce fine output details. The number of samples is arguably the most important parameter for any NeRF model because it is directly related to the training time, the prediction time, and the quality of the novel view. The following experiments in Table~\ref{table:samplenum} demonstrate that the proposed method performs significantly well even when the number of samples is low. The 64 samples provide the best compromise between novel view quality and training time. 
\begin{table}[htb]
\small
	\centering
	\begin{tabular}{l|lll}
	\hline
     & \multicolumn{3}{|c}{Number of samples} \\
	\hline
	Metrics & 16 & 64 & 128 \\
	\hline
	PSNR $\uparrow$ & 19.4  & \cellcolor[HTML]{34FF34}21.18 & \cellcolor[HTML]{32CB00}22.13  \\
	
	SSIM $\uparrow$ & 0.86 & \cellcolor[HTML]{34FF34}0.89  & \cellcolor[HTML]{32CB00}0.9 \\
	
	Abs Rel $\downarrow$ & 0.05 & \cellcolor[HTML]{32CB00}0.04 & 0.05\\
	
	LPIPS $\downarrow$ & 0.0025 & 0.0023 & \cellcolor[HTML]{32CB00}0.0018\\
	
	Training Time $\downarrow$ & \cellcolor[HTML]{32CB00}38m & \cellcolor[HTML]{34FF34}44m & \cellcolor[HTML]{CB0000}1.54h \\
	
	\end{tabular}
	
	\caption{A comparison between training time and novel view quality based on the number of samples per ray.}
	\label{table:samplenum}
\end{table}

\subsubsection{Different datasets}
The proposed method was evaluated with 4 different datasets with different characteristics to demonstrate its robustness in different types of scenes. The cube has a simple geometry but a complicated texture. Alternatively, the drums have a complicated and very detailed 3D structure. The Lego scene is a good mix of photometric and geometric details. The human scene mimics some real-world applications. Quantitative results are shown in Table~\ref{table:datasets} and qualitative results are shown in Figure~\ref{fig:lcd}.

\begin{figure*}[ht!]
\begin{center}
   \includegraphics[width=0.75\linewidth]{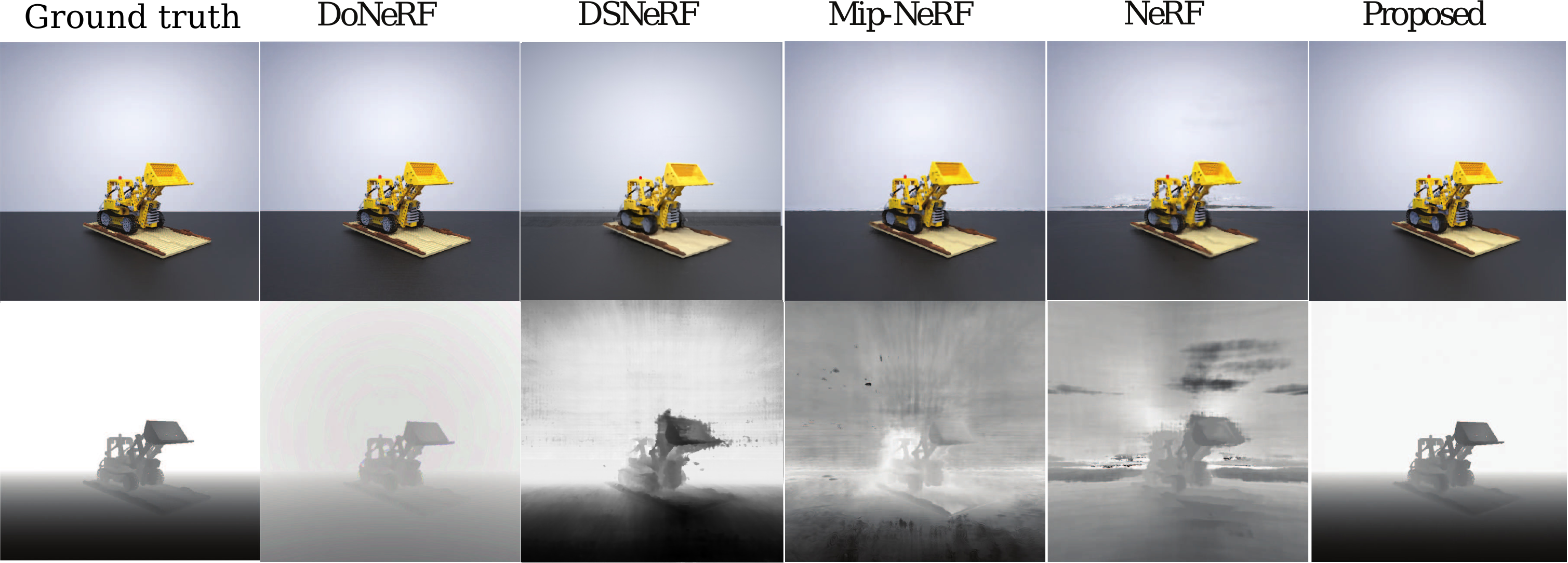}
\end{center}
   \caption{Qualitative comparison: Visual comparison results between the proposed method and other state-of-the-art methods.}
\label{fig:compare}
\end{figure*}
\begin{figure*}[ht!]
\begin{center}
   \includegraphics[width=0.5\linewidth]{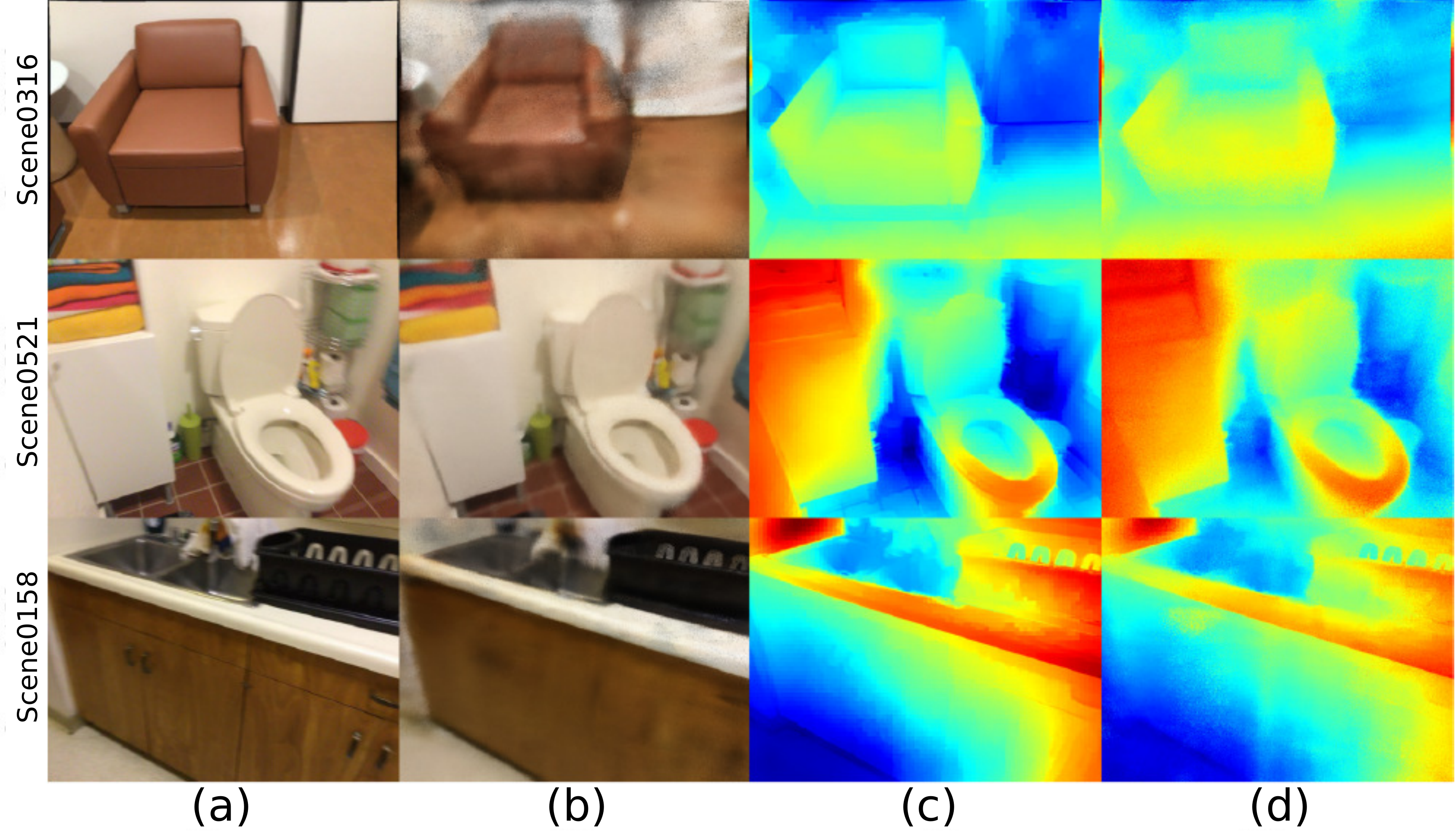}
\end{center}
   \caption{Qualitative result of the proposed method on 3 different real sensor datasets. (a) Ground truth RGB images; (b) Predicted RGB images; (c) Ground-truth depth maps; (d) Predicted depth maps.}
\label{fig:mvsdataset}
\end{figure*}

\begin{table}[htb]
\small
	\centering
	\begin{tabular}{p{1.4cm}|p{1cm}p{1cm}p{1cm}p{1.1cm}}
	\hline
      & \multicolumn{4}{|c}{Metrics} \\
    \hline
    dataset & PSNR$\uparrow$ & SSIM$\uparrow$ & AbsRel$\downarrow$ & LPIPS$\downarrow$ \\ 
	\hline
	Lego & 28.21 & 0.93 & 0.02 & 0.0013 \\
	
	Cube & 22.78 & 0.95 & 0.01 & 0.0001 \\
	
	Human & 37.7 & 0.97 & 0.02 & 0.00008 \\
	
	Drums & 27.85 & 0.9 & 0.02 & 0.0012 \\
	\end{tabular}
	
	\caption{Performance of the proposed method on 4 different simulated datasets. }
	\label{table:datasets}
\end{table}

\subsubsection{Fewer input views}
To demonstrate that the proposed method can perform well even when the number of training images is limited, three different datasets with different numbers of training images have been considered for experiments. Table~\ref{table:numinput} shows a comparison between different datasets. Although increasing the number of inputs increases the quality of the novel view, the training time also increases significantly.  

\begin{table}[htb]
	\centering
	\begin{tabular}{l|lll}
	\hline
     & \multicolumn{3}{|c}{Number of input views} \\
	\hline
	Metrics & 8 & 30 & 100 \\
	\hline
	PSNR $\uparrow$ & 21.18  &  25.25  & \cellcolor[HTML]{32CB00}28.21 \\
	
	SSIM $\uparrow$ & 0.89 & 0.92 & \cellcolor[HTML]{32CB00}0.93 \\
	
	Abs Rel $\downarrow$ & 0.04 & 0.04 & \cellcolor[HTML]{32CB00}0.02\\
	
	LPIPS $\downarrow$ & 0.0023 & 0.0017 & \cellcolor[HTML]{32CB00}0.0013 \\
	
	Training Time $\downarrow$ & \cellcolor[HTML]{32CB00}44m & 2.15h & \cellcolor[HTML]{CB0000}6.45h\\
	\end{tabular}
	\caption{More samples in the training set provides more supervision for the network to learn the scene representation. With an increasing number of input views, geometric and photometric metrics improve. Subsequently, training time increases significantly. A dataset with 100 images takes 11 times longer to train than dataset with 8 images.}
	\label{table:numinput}
\end{table}
\subsubsection{Scene representation}\label{comparenerf}
\begin{table}[htb]
\small
	\centering
	\begin{tabular}{p{1.3cm}|p{0.7cm} p{0.7cm} p{0.7cm} p{0.8cm} p{0.7cm}}
	\hline
      & \multicolumn{5}{|c}{Metrics} \\
    \hline
    Method & PSNR $\uparrow$ & SSIM $\uparrow$ & Abs Rel$\downarrow$ & LPIPS $\downarrow$ & Time $\downarrow$ \\ 
	\hline
	DSNeRF & 29.31 & 0.87 & 0.489 & 0.003 & 3:37h\\
	
	DONeRF & \cellcolor[HTML]{32CB00}39.23 & \cellcolor[HTML]{32CB00}0.98 & 0.008 & \cellcolor[HTML]{32CB00}0.00001 & \cellcolor[HTML]{CB0000}5:08h\\
	
	NeRF & 27.36 & 0.94 & 0.34 & 0.0015 & 3.40h\\
	
	MipNeRF & 30.66 & 0.95 & 0.334 & 0.006 & 1:27h\\
	
	\textbf{Proposed} & \cellcolor[HTML]{34FF34}32.72 & \cellcolor[HTML]{34FF34}0.95 & \cellcolor[HTML]{32CB00}0.001 & \cellcolor[HTML]{34FF34}0.0004 &
	\cellcolor[HTML]{32CB00}1.15h\\
	\end{tabular}
	\caption{Quantitative comparison for novel view synthesis and depth estimation between the proposed method and state-of-the-art methods. The Lego dataset is used for all these experiments.}
	\label{table:compare}
\end{table}

\begin{table}[htb]
\small
	\centering
	\begin{tabular}{p{1.4cm}|p{1cm}p{1cm}p{1cm}p{1.1cm}}
	\hline
      & \multicolumn{4}{|c}{Metrics} \\
    \hline
    dataset & PSNR$\uparrow$ & SSIM$\uparrow$ & AbsRel$\downarrow$ & LPIPS$\downarrow$ \\ 
	\hline
	scene0521 & 25.48 & 0.724  & 0.025 & 0.0004 \\
	
	scene0316 & 16.99 & 0.57 & 0.05 & 0.001 \\
	
	scene0158 & 24.93 & 0.74 & 0.02 & 0.0007 \\
	\end{tabular}
	
	\caption{Performance of the proposed method on 4 different real RGB-D datasets. }
	\label{table:realdataset}
\end{table}

In this section, the proposed method is compared with other state-of-the-art NeRF-based methods. 

\textbf{NeRF} \cite{mildenhall2020nerf}: The implementation of PyTorch Lighting of the NeRF by \cite{queianchennerf} has been considered for the experiments. NeRF can be trained using simulated 360-degree Blender data or real data. For these particular experiments, simulated Blender scenes were used. 

\textbf{DSNeRF} \cite{kangle2021dsnerf}: DSNeRF works only on the forward facing scenes where depth supervision data is generated using Colmap \cite{schonberger2016structure}. The official implementation of DSNeRF was used for these experiments. 

\textbf{DONeRF} \cite{neff2021donerf}: DONeRF works only on forward-facing datasets where all poses belong to a view cell. This method works only with simulated data with a dense depth map. The official implementation of the DONeRF was used for the experiments. 

\textbf{Mip-NeRF} \cite{barron2021mip}: The official Mip-NeRF implementation on JAX was converted to PyTorch for convenience of comparison.  

All experiments were carried out on the same Lego scene dataset that contains 30 training images with resolution $800 \times 800$. The quantitative results in Table: \ref{table:compare} and the qualitative results in Figure: \ref{fig:compare} show that DONeRF \cite{neff2021donerf} can produce the best photometric quality, but is limited to forward-facing scenes, a longer training time, and oracle network-based depth prediction, where the proposed method uses only one smaller network (less space requirement), trains faster (4 times faster), and produces more accurate geometry. 

Three different real-world datasets have been used from the NerfingMVS~\cite{wei2021nerfingmvs} for experimenting on real acquired depth images. Pre-processed depth maps were used instead of using raw depth maps because the raw depth maps contain areas without depth information (holes where sensors cannot estimate depth). NerfingMVS uses a monocular depth prediction network to complete the missing depths. Alternatively, holes in the raw depth maps could be handled as in the classic RGB NeRF implementation, however, this randomly affects computational performance and comparisons. Table:\ref{table:realdataset} shows the quantitative results of the proposed method on 3 different datasets. The adaptive sampling strategy with 16 samples was used for all of these experiments. The qualitative results of the experiments are shown in the Figure:\ref{fig:mvsdataset}. The ground-truth depth shows that it is not very detailed and that some areas have wrong depth measurements, which results in some artifacts in the predicted image generated by the proposed method.  
	
	
	

\subsection{Analysis}
\textit{Fewer views:} The proposed method can learn a scene representation from fewer views, as depth supervision provides additional supervision and effective sampling. Depth supervsion allows the network to learn scene geometry and multi-view constancy from a very limited number of views.  

\textit{Faster training:} The results show a quantifiable speed improvement in training time with the proposed method compared to other state-of-the-art methods. Faster training was achieved using fewer samples, a smaller network architecture, and local sampling. The Mip-NeRF RGB-D method is $3-5\times$ faster compared to other similar NeRF-based methods.

\textit{Accurate depth estimation:} The proposed method is capable of producing a more accurate geometry compared to other state-of-the-art methods. The network can learn accurate geometry from small number of inputs as few as 8 frames. 

\section{Discussion}
In this article, a new method was presented for representing 3D scenes from RGB-D data using recent neural radiance fields. Instead of learning the radiance field from RGB images, the proposed method uses RGB-D frames, which allows achieving better underlying geometry and faster training and prediction times.  Additional depth supervision of dense depth maps is shown to have a significant improvement on the training time through local sampling. The proposed method trains $3-5\times$ faster and improves the novel view and depth quality. The experiments show significant improvements over the state-of-the-art methods, both quantitatively and qualitatively. 
Future perspective will be focused on extending this approach to dynamic scenes. 

\section{Acknowledgements}
This project has received funding from the H2020 COFUND
program BoostUrCareer under Marie Sklodowska-Curie
grant agreement no. 847581. It also received funding from the EU H2020 MEMEX research project under grant agreement No. 870743. This work was granted access to the HPC resources of IDRIS under the allocation 2021-AD011012578 made by GENCI.

%% file: method.tex
Now the proposed Mip-NeRF RGB-D will be presented. First, the implicit scene representation used by NeRF based methods will be over-viewed, followed by the explanation of the rendering process. After that, an efficient network architecture will be proposed and a joint optimization method using RGB-D data will be presented. Finally, the local sampling strategy used to reduce the number of samples along the rays and reduce training time will be described. 
\subsection{Implicit scene representation}
The proposed system is based on the Mip-NeRF \cite{barron2021mip} method which is an extension of NeRF for handling anti-aliasing. Vanilla NeRF based methods use a set of images and corresponding poses to train a MLP network that represents the scene by outputting the emitted radiance and volume density of 3D locations. Given 5D coordinates(3D location + viewing direction) as input, the network \(F_\Theta\) learns an implicit function that estimates color \(C=(r,g,b)\) and volume density \(\tau\) as: \begin{equation}
    F_\Theta : (x, y, z, \theta, \phi ) \rightarrow ( C, \tau).
\end{equation}

First, rays \(r(t) = \textbf{o} + t\textbf{d}\) passing through each pixel of the image are generated, where the ray origin \(\textbf{o}\) is the camera center and \(\textbf{d}\) is the ray direction. Then N sample points are placed along the ray stratified manner between predefined near and far bounds. The color of each pixel is computed using a radiance and a volume density along the ray. In Mip-NeRF the rays are replaced with cones generated using the camera center and the pixel size. The cone is split into N intervals \(\mathcal{T_i} = [t_i, t_i+1)\) and for each interval the integrated positional encoding of the mean and the covariance \((\boldsymbol{\mu},\Sigma)\) of the corresponding conical frustum is computed. Integrated positional encoding encodes the Gaussian approximation of the conical frustum as follows:
\begin{equation}
\gamma(\boldsymbol{\mu},\Sigma)=\begin{Bmatrix}\begin{bmatrix}\sin{(2^l \mu)}exp(-2^{l-1}diag(\Sigma)) \\
\cos{(2^l \boldsymbol{\mu})}exp(-2^{l-1}diag(\Sigma))\end{bmatrix}\end{Bmatrix}^{L-1}_{l=0},
\end{equation} where \(\Sigma\) is the covariance of the Gaussian approximation: \begin{equation}
\Sigma = \sigma^2_t(\textbf{d} \cdot \textbf{d}^\mathrm{T}) + \sigma^2_r\Big(I - \frac{\textbf{d} \cdot \textbf{d}^\mathrm{T}}{||\textbf{d}||^2_2}\Big).    
\end{equation} 

The variance along the ray is denoted by \(\sigma_t^2\) and the variance perpendicular to the ray is \(\sigma_r^2\).
Mip-NeRF uses this integrated positional encoding instead of the frequency positional encoding as input to the neural network. One of the key difference between Mip-NeRF and the proposed method is the local sampling strategies, which will be discussed in the subsequent part of this article.

\subsection{Volume rendering}
Similarly to NeRF, a volume rendering formula was used to calculate the color and the depth of pixels from radiance and volume density of the conical frustum. The volume density \(\tau(P)\) at location \(P=(x,y,z)\) can be interpreted as the differential probability of ray termination. The expected color \(C(r)\) of a camera ray \(r(t) = \textbf{o} + t\textbf{d}\) with near and far bounds \(t_n\) and \(t_f\) is: 
\begin{equation}
    C(r) = \int_{t_n}^{t_f} T(t)\tau(r(t))c(r(t),\textbf{d})dt,
\end{equation} where \begin{equation}
    T(t)=exp(-\int_{t_n}^{t} \tau(r(s))ds).
\end{equation}

The function \(T(t)\) denotes the accumulated transmittance along the ray from \(t_n\) to \(t\), i.e., the probability that the ray travels from \(t_n\) to t without hitting any other particle. In the stratified sampling approach \([t_n, t_f ]\) is partitioned into $N$ evenly-spaced bins and then one sample is drawn uniformly at random from within each bin. The samples are used to estimate predicted color $\widehat{C}(r)$ as:

\begin{equation}\widehat{C}(r) = \sum_{i=1}^{N}T_i(1-exp(-\tau_i\delta_i))c_i,\end{equation} where \begin{equation}
    T_i = exp(- \sum_{j=1}^{i-1}\tau_j\delta_j).
\end{equation}

Here \(\delta_j=t_{j+1}-t_j\) is the distance between adjacent samples. Similarly \cite{wei2021nerfingmvs}, the depth can be represented with volume density using: \begin{equation}
    \widehat{D}(r)=\sum_{i=1}^{N}T_i(1-exp(-\tau_i\delta_i))t_i,
\end{equation} where \(T_i\) is the accumulated transmittance. 
To optimize the network, NeRF uses a squared error between the rendered and true pixel colors.

\subsection{Optimization}
The network parameters \(\theta\) are optimized using a set of RGB-D frames, each of which has a color, depth, and camera pose information. The proposed method minimizes the geometric and photometric loss together on a set of frames as the rendering functions are completely differentiable. The photometric loss $l_p$ is the absolute difference (L1-norm)\cite{sucar2021imap} between the predicted color and the ground truth color of the ray. The photometric loss over a set of rays is defined as: \begin{equation}
    l_p= \sum_{r \in R}\lvert \hat{C}(r) -C(r) \rvert.
\end{equation}

The geometric loss is the absolute difference between predicted and true depths, normalized by the depth variance \cite{sucar2021imap} to discourage weights with high uncertainty:
\begin{equation}
    l_g = \sum_{r \in R}\frac{\lvert \hat{D}(r)-D(r) \rvert}{\sqrt{\hat{D}_{var}(r)}},
\end{equation} where \(\hat{D}_{var}(r) = \sum_{i=1}^{N} T_i(1 - exp(-\tau_i\delta_i))(\hat{D}(r) - t_i)^2\) depth variance of the image.
The neural network can be optimized by combining photometric, and geometric losses together using  empirically chosen scale factors~\(\lambda_p\): \begin{equation}
    min_{\theta}(l_g + \lambda_p l_p).
\end{equation}

\subsection{Network architecture}
The network architecture is similar to the original NeRF with some modifications. The proposed method uses only one network with 4 hidden layers of feature size 256. The skip connection is used in layer 3. The viewing direction is concatenated to the fourth layer before the color and volume density are output. The integrated positional encoding was applied to the conical frustum and a positional encoding of frequency 4 is applied to the viewing directions as was done in Mip-NeRF. By decreasing the network size, faster training and prediction time were achieved without significantly compromising the novel view quality. 

\subsection{Local sampling}
\begin{figure}[t]
\begin{center}
   \includegraphics[width=1\linewidth]{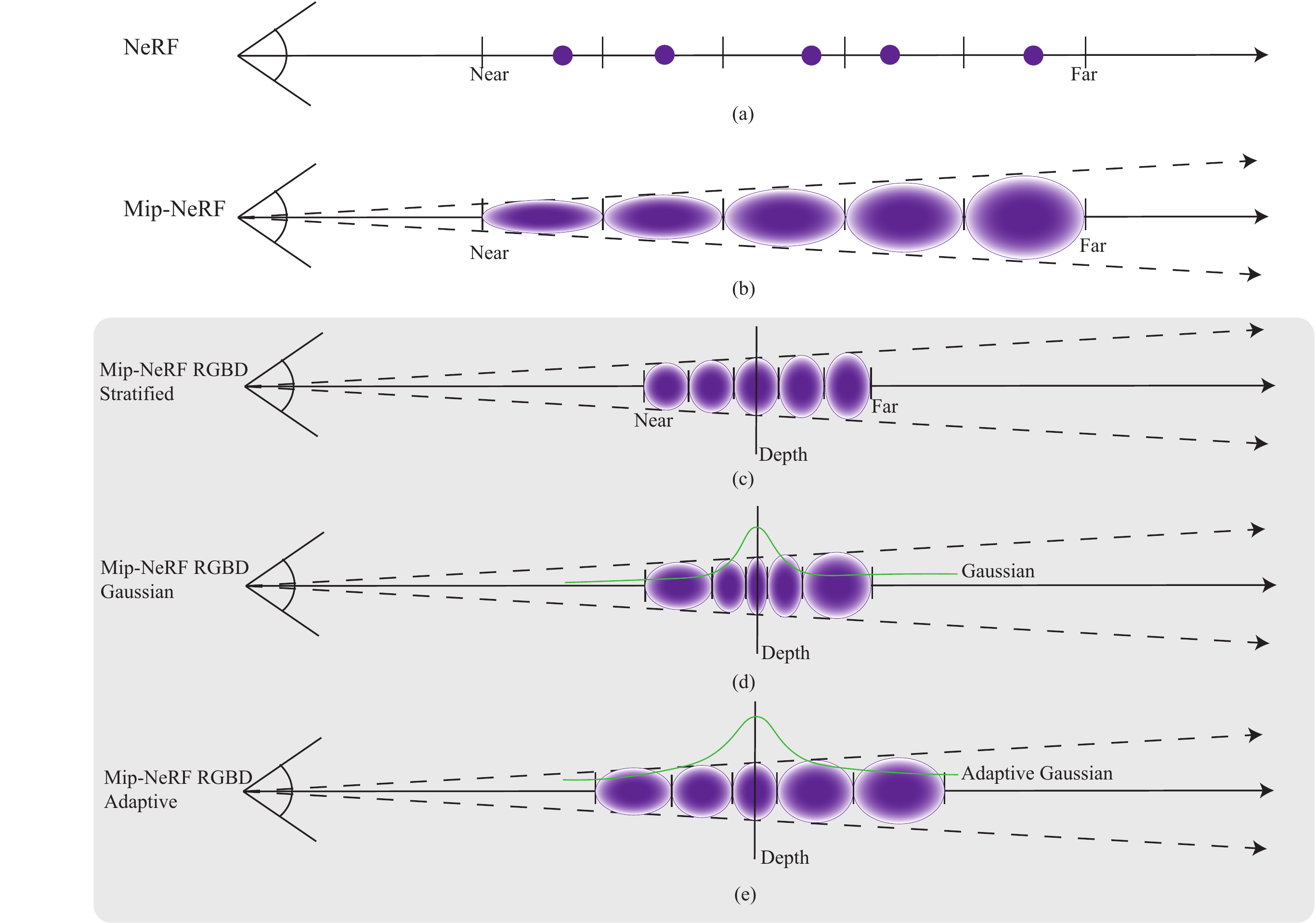}
\end{center}
   \caption{Visualization of the proposed sampling strategies(c, d, e) compared to NeRF(a) and Mip-NeRF(b). Black arrows represent ray direction and purple ellipsoids represent a Gaussian approximation of the conical frustum.}
\label{fig:sampling}
\end{figure}

NeRF based methods estimate pixel color by placing samples on viewing rays traced through the pixels. The final color of the pixels is calculated by the alpha composition \cite{max1995optical} of the volume density and the radiances of the samples along the ray. Samples relevant to the volume produce higher volume density, so samples close to the surface are more relevant to the pixel's final color. NeRF uses 256 samples in a stratified manner and 2 networks to ensure that samples are placed on relevant parts of the ray. To compute a pixel color, each sample on that ray needs a full network evaluation, so the training time increases exponentially with the number of samples on the ray. Although Mip-NeRF uses a conical frustum instead of viewing rays, it still requires 2 network passes and a large number of bins to create integrated positional encoding. In reality, the majority of the scene volume is empty space (for 360 scenes), and the samples placed on the empty space have less contribution to the final color. Therefore, given the depth information of an image, it is possible to place fewer samples and to place them directly on the relevant parts of the ray, while achieving similar quality results. Finally, in this case, it is also possible to eliminate the coarse network with local sampling, which NeRF uses to find important sampling locations along the ray. Various depth-guided sampling strategies have been considered. Figure:\ref{fig:sampling} shows the comparison between the proposed local sampling and the baseline approaches.

\subsubsection{\emph{Stratified sampling}}
The so-called stratified sampling strategy is very similar to the original Mip-NeRF sampling, but the conical frustums are generated only close to the surface based on depth information(Figure:\ref{fig:sampling}(c)). Here the space between the near and far bounds $[t_n,t_f]$ is divided into $N$ evenly spaced bins and a sample is drawn uniformly at random from each bin where $t_n = D - \alpha_n$ and $t_f = D + \alpha_f$, $\alpha_n$ and $\alpha_f$ are empirically chosen based on the depth uncertainty. The samples are then used as the bounds of the conical frustum. This allows the network to avoid empty space and eventually decrease the number of bins needed for each ray.  

\subsubsection{\emph{Gaussian sampling}}
In this strategy, instead of placing the bins equidistantly around the surface, the limits of the conical segments are selected from a normal distribution where the mean is the depth and the standard deviation \(\varsigma\) is empirically chosen based on the depth uncertainty. In this way, it ensured that the conical frustums are smaller (toward the ray direction) on the relevant part of the ray(close to the true depth as in Figure:\ref{fig:sampling}(d)) to emphasize the high-frequency details on the surface. This allows the network to handle the generalized uncertainty present in the depth estimate. 
\subsubsection{\emph{Adaptive sampling}}

The adaptive sampling strategy uses a normal distribution with a varying standard deviation \( \varsigma(r)\) based on the number of epochs and the depth of the ray. Therefore, the normal distribution (the mean is the depth measurement) is used to define the limits of the conical frustums(Figure:\ref{fig:sampling}(e)). \(\varsigma(r)\) varies during training according to the number of epochs in a coarse to fine manner to improve the fine photometric details. Additionally, this sampling strategy takes into account the depth uncertainty, which increases with distance. The standard deviation of each ray is calculated as follows:

\begin{equation}
    \varsigma(r) = \frac{D(r)}{4} ( \exp^{- \lambda_r i} + \lambda_m),
\end{equation}
where \(i\) is the epoch number, \(\lambda_r\) is the rate of decrease, \(\lambda_m\) is minimum standard deviation, and \(D(r)\) is the true depth of the ray. \(\lambda_r\) and \(\lambda_m\) are empirically chosen based on dataset and depth uncertainty.